\documentclass{article}
\usepackage{ijcai23}

\usepackage{times}
\usepackage{soul}
\usepackage{xspace}
\usepackage{url}
\usepackage[hidelinks]{hyperref}
\usepackage[utf8]{inputenc}
\usepackage[small]{caption}
\usepackage{graphicx}
\usepackage{amsmath}
\usepackage{amssymb}
\usepackage{amsthm}
\usepackage{xcolor}
\usepackage{booktabs}
\usepackage{multirow}
\usepackage{algorithm}
\usepackage{algorithmic}
\usepackage[switch]{lineno}
\usepackage[capitalize]{cleveref}
\usepackage{color, colortbl}
\crefname{section}{Sec.}{Secs.}
\Crefname{section}{Section}{Sections}
\Crefname{table}{Table}{Tables}
\crefname{table}{Table}{Tables}
	
\definecolor{LightCyan}{rgb}{0.88,1,1}

\def\R{\mathbb{R}}
\newcommand{\best}[1]{\textcolor{red}{\textbf{#1}}}
\newcommand{\sbest}[1]{\textcolor{blue}{\textit{\underline{#1}}}}

\makeatletter
\DeclareRobustCommand\onedot{\futurelet\@let@token\bmv@onedotaux}
\def\bmv@onedotaux{\ifx\@let@token.\else.\null\fi\xspace}
\def\eg{\emph{e.g}\onedot} 
\def\ie{\emph{i.e}\onedot} 
 
\def\etc{\emph{etc}\onedot}

\def\bigoh{\mathcal{O}}
\makeatother

\title{Diagnose Like a Pathologist: Transformer-Enabled Hierarchical Attention-Guided Multiple Instance Learning for Whole Slide Image Classification}

\author{
Conghao Xiong$^1$\and
Hao Chen$^2$\and
Joseph J.Y. Sung$^3$\And
Irwin King$^1$
\affiliations
$^1$Department of Computer Science and Engineering, The Chinese University of Hong Kong\\
$^2$Department of Computer Science and Engineering and Department of Chemical and Biological Engineering, The Hong Kong University of Science and Technology\\
$^3$Lee Kong Chian School of Medicine, Nanyang Technological University
\emails
\{chxiong21, king\}@cse.cuhk.edu.hk,
jhc@cse.ust.hk,
josephsung@ntu.edu.sg
}

\begin{document}

\maketitle

\begin{abstract}
Multiple Instance Learning (MIL) and transformers are increasingly popular in histopathology Whole Slide Image (WSI) classification. However, unlike human pathologists who selectively observe specific regions of histopathology tissues under different magnifications, most methods do not incorporate multiple resolutions of the WSIs, hierarchically and attentively, thereby leading to a loss of focus on the WSIs and information from other resolutions. To resolve this issue, we propose a Hierarchical Attention-Guided Multiple Instance Learning framework to fully exploit the WSIs. This framework can dynamically and attentively discover the discriminative regions across multiple resolutions of the WSIs. Within this framework, an Integrated Attention Transformer is proposed to further enhance the performance of the transformer and obtain a more holistic WSI (bag) representation. This transformer consists of multiple Integrated Attention Modules, which is the combination of a transformer layer and an aggregation module that produces a bag representation based on every instance representation in that bag. The experimental results show that our method achieved state-of-the-art performances on multiple datasets, including Camelyon16, TCGA-RCC, TCGA-NSCLC, and an in-house IMGC dataset. The code is available at \url{https://github.com/BearCleverProud/HAG-MIL}.
\end{abstract}

\section{Introduction}
Histopathology tissue analysis is widely recognised as the gold standard for cancer diagnosis, making histopathology \textbf{W}hole \textbf{S}lide \textbf{I}mage (WSI) analysis an important deep learning application in the real world. Due to the advancements of scanning and storage devices, the number of digitalised histology tissues, \ie, WSIs, is increasing. This makes deep learning methods applicable to WSI classification tasks \cite{dimitriou_survey_2019,srinidhi_survey_2021,li_comprehensive_2022}, and there are already some successful applications on breast \cite{maximizing_huang,peng_robust_2008}, prostate \cite{campanella_clinical-grade_2019}, skin \cite{ianni_tailored_2020}, and pancreas cancer \cite{keikhosravi_non-disruptive_2020}.
\begin{figure}
    \centering
    \includegraphics[width=\linewidth]{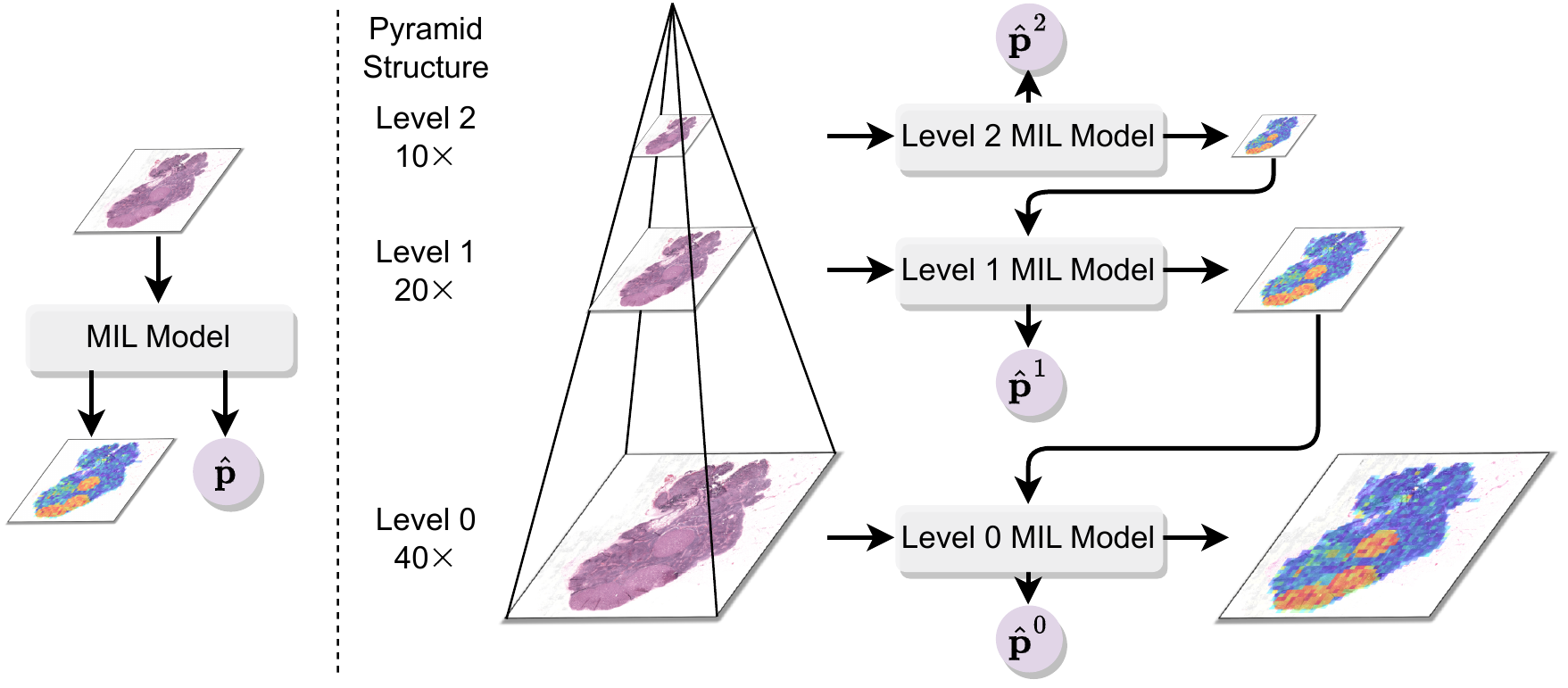}
    \caption{An illustration of the difference between AB-MIL and the HAG-MIL framework. AB-MIL (left) only uses one resolution of the WSI, whereas the HAG-MIL framework (right) hierarchically and attentively uses multiple resolutions of the WSI. The high-attention regions are passed to the models at higher resolutions for further checking.}
    \label{fig:HAGMIL_illustration}
\end{figure}

Though deep learning is a common solution to computer vision \cite{resnet_he_16,swin_transformer}, WSIs, as gigapixel images, pose a unique challenge for deep learning since they are too large to be fed into a neural network. The \textbf{M}ultiple \textbf{I}nstance \textbf{L}earning (MIL) \cite{NIPS1997_MIL} framework provides a solution to this problem. Under this framework, the WSI is regarded as a bag of patches (instances) that are images cropped from the WSI, which can be directly fed into a neural network \cite{ilse_attention-based_2018}.

Most MIL methods assume that the patches are independent and identically distributed \cite{ilse_attention-based_2018,campanella_clinical-grade_2019}. However, patches have strong correlations with each other, \eg, a tumour patch might have similar morphological structures to another tumour patch in the same WSI. Neglecting these correlations may result in insufficient exploitation of the WSI. The transformer \cite{transformer_2017} is well-suited for MIL with patch correlations, as it can capture the dependencies between patches \cite{shao_transmil}.

The lack of fine-grained annotations is another obstacle to achieving good performance with deep learning models. Due to the large size of the WSI, it can take several hours for an expert pathologist to annotate each pixel, which is prohibitively time-consuming and expensive; therefore, the WSI datasets usually include only slide-level labels. Furthermore, the malignant tissues may only occupy less than 10\% of the entire WSI \cite{litjens_1399_2018}, making most of the regions irrelevant to the bag label ``tumour''. It is therefore important to find the discriminative patches from WSIs since they are the only patches providing strong indications of the bag label. These factors also contribute to the thriving of the weakly-supervised learning on WSI classification tasks \cite{lu_semi-supervised_2019,chikontwe_multiple_2020,Zhao_2020_CVPR,lu_data-efficient_2021,tourniaire_attention-based_2021,li_2021_dual,li_multi_resolution_2021}, as they do not need pixel-level annotations.

Identifying the discriminative patches in a large dataset can be a challenging task, akin to finding a needle in a haystack. The WSI datasets often have less than a thousand of WSIs whereas hundreds of thousands of patches can be cropped from each WSI. These discriminative patches are overwhelmed by a flood of irrelevant ones, preventing the model from learning the most essential features of the patches. In fact, many works address this issue through techniques like contrastive learning \cite{li_2021_dual,transpath_2021,tu_dccml}, attention mechanism \cite{ilse_attention-based_2018,tourniaire_attention-based_2021,lu_data-efficient_2021,li_multi_resolution_2021}, hard mining \cite{li_deep_2019}, and so on \cite{wang_weakly_2020}.

We address the problem of finding the most discriminative patches from the following two aspects.

\textbf{Multi-resolution} Previous methods for WSI classification often attach little importance to utilisation of multiple resolution explicitly \cite{scaling_chen_2022}. For instance, a patch at 40$\times$ magnification (level 0) may contain fine-grained cellular features, such as stroma, tumour cells, lymphocytes and other texture and morphology features. However, a patch at 10$\times$ magnification (level 2) may have the tissue microenvironment, including the organisations of the cell structures and interactions between clusters of cells and so on. Using only one resolution leads to a lack of information and multiple resolutions should be exploited. Unlike current methods, human pathologists intentionally inspect histopathology tissues under multiple magnifications, switching their focus on different regions and magnifications of the WSIs based on their prior knowledge and observations. They diagnose those in a hierarchical and attentive way, which is not the case with current model behaviours. Therefore, to incorporate multiple resolutions of the WSIs and from being inspired by the human pathologists, we propose a \textbf{H}ierarchical \textbf{A}ttention-\textbf{G}uided \textbf{M}ultiple \textbf{I}nstance \textbf{L}earning (HAG-MIL) framework to minimise the disparity between model and human behaviours. 

\Cref{fig:HAGMIL_illustration} illustrates how human pathologists make diagnoses and how our framework is designed to resemble their approach. Information from lower resolutions can guide the pathologists to regions of interest at higher resolutions for further refinement of their judgment. Similarly, the HAG-MIL framework first inspects patches from lower resolutions, discovers the suspicious regions and passes them to another model at higher resolutions for further confirmation. The process is repeated until the highest resolution is reached. By reserving only a certain number of patches at each resolution, the number of patches in a bag is maintained below a specified bag size. This enables the models to operate at the highest resolution of the WSIs. Also, by eliminating irrelevant patches, the models can concentrate more on the most discriminative ones, thereby improving their performance.

\textbf{Structure Flaws in the Transformer Structure} Although transformers are popular models in all fields, they only output one bag representation at the last layer using the class token \cite{devlin-etal-2019-bert,shao_transmil,scaling_chen_2022}. This structure has two shortcomings: (1) the bag representation should be obtained based on all instance representations, not just the class token and (2) the bag representation can evolve through different transformer layers and relying solely on the bag representation from the last layer may result in a loss of information from previous bag representations. Hence, we propose the \textbf{I}ntegrated \textbf{A}ttention \textbf{T}ransformer (IAT), consisting of multiple stacked \textbf{I}ntegrated \textbf{A}ttention \textbf{M}odules (IAMs). This module uses an aggregation module to aggregate all instance representations into a bag representation, instead of relying on the class token. Multiple bag representations from different modules are fused into a final bag representation, which allows the transformer model to take into account the evolution of bag representations and produce a more effective and holistic bag representation. 

In summary, the benefits of our method include: (1) the alignment of our method with the visual assessment of human pathologists by its effective utilisation of multiple resolutions of the WSIs, (2) preventing discriminative patches from being overwhelmed by irrelevant ones by discarding certain amount of irrelevant patches from the WSI bags, (3) enabling the models performing at the highest resolutions, and (4) generating a more holistic and effective bag representation for the WSI by taking into account all instance representations in that bag and the evolution of the bag representation through different layers. 
Our contributions are summarised below:
\begin{enumerate}
    \item We present the HAG-MIL framework to dynamically, attentively and hierarchically locate the most discriminative patches across multiple resolutions of the WSIs like a human pathologist.
    \item We propose the IAT to produce a more effective and holistic bag representation by generating a bag representation in each IAM based on every instance representation, and then fusing them together.
    \item We conduct extensive experiments on various datasets including Camelyon16, TCGA-RCC, TCGA-NSCLC, and our in-house IMGC dataset. The results demonstrate that our method consistently outperforms other methods.
\end{enumerate}

\section{Related Work}
\subsection{MIL in WSI Analysis}
MIL can be divided into two approaches: (1) the \textit{instance-level} and (2) the \textit{embedding-level} approaches. In the first approach, a classifier is trained at the instance level using bag labels, and the predicted instance labels are then aggregated to form the bag label prediction \cite{hou_patch_based_2016,campanella_clinical-grade_2019,kanavati_weakly-supervised_2020,chikontwe_multiple_2020}. In the second approach, a bag representation is first obtained from instance representations, and a classifier is trained on the bag representations \cite{ilse_attention-based_2018,hashimoto_multi-scale_2020,li_deep_2020,tourniaire_attention-based_2021,lu_data-efficient_2021,li_2021_dual,shao_transmil,zhang_dtfd-mil_2022}. The second approach is shown to be superior to the first one, as the instance labels tend to be noisy \cite{WANG_MIL}; therefore, in our work, we adopt the second approach.

Despite its effectiveness, MIL has an apparent disadvantage that it treats every patch in a bag as equally important. Patches should be assigned different importance scores so that the discriminative patches have high importance scores. \textbf{A}ttention-\textbf{B}ased \textbf{MIL} (AB-MIL) \cite{ilse_attention-based_2018} calculates the attention scores from instance representations and the attention scores reflect the importance of the corresponding patches. In this way, the model obtains the bag representation by taking a weighed average of the instance representations. Many recent works adapt the idea from AB-MIL \cite{li_2021_dual,lu_data-efficient_2021,zhang_dtfd-mil_2022}.

\subsection{Transformers in WSI Analysis}
Transformers \cite{transformer_2017,swin_transformer} have demonstrated their efficacy due to their ability to capture long-range dependencies. There are also many works on WSI classification tasks that utilise transformers to model correlations between patches \cite{shao_transmil,integration_transformer_2021,accounting_2021,multimodal_transformer_2021,graph_transformer_2022,scaling_chen_2022}. In our work, we further improve the quality of the bag representation using an aggregation module, which produces a bag representation based on all instance representations, and bag representations from different IAMs are then fused into a holistic and effective bag representation.

\subsection{Utilising Multiple Resolutions of the WSIs}
Pathologists make diagnoses with multiple magnifications of histopathology tissues, and it is useful to utilise multiple resolutions of the WSIs \cite{babak_2015,tokunaga_adaptive_2019,hashimoto_multi-scale_2020,li_2021_dual,zhou_2022_hierarchical}. However, the previous works only used multiple resolutions in the same bag \cite{hashimoto_multi-scale_2020} or concatenate the features from different resolutions together \cite{li_2021_dual}. 
Moreover, the two-stage training can be reduced to an end-to-end process \cite{zhou_2022_hierarchical}.
Different from other works, we resemble the pathologists and make use of multiple resolutions in a hierarchical and attentive way.

\section{Methodology}
\begin{figure*}
    \centering
    \includegraphics[width=\linewidth]{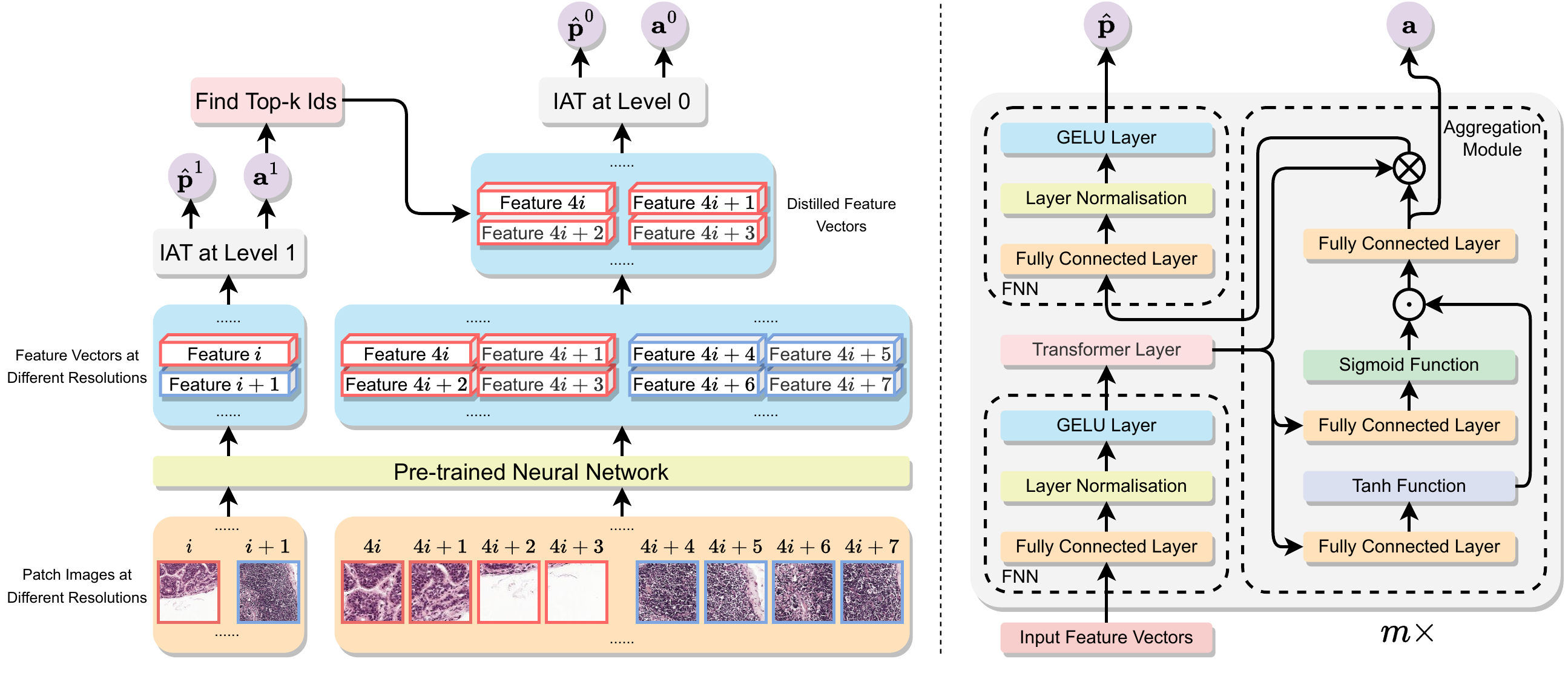}
    \caption{Overview of the HAG-MIL framework (left) and the model architecture of our proposed transformer model IAT (right). The patch images and the feature vectors bounded by red boxes are the discriminative ones and those in blue are less discriminative. The patch images cropped from different resolutions of WSIs are fed into a pre-trained network to obtain the feature vectors. The feature vectors are then passed to the framework in a hierarchical way to obtain the attention score $\boldsymbol{a}^1$ and $\boldsymbol{a}^0$, and the probability distribution of the bag label $\hat{\boldsymbol{p}}^1$ and $\hat{\boldsymbol{p}}^0$.}
    \label{fig:archi}
\end{figure*}
\subsection{Problem Formulation}
Given the features extracted by a pre-trained neural network $\boldsymbol{F} \in \R^{n\times d}$, where $n$ is the number of patches in the WSI and $d$ is the hidden dimension of the pre-trained neural network, the MIL model predicts the bag label of that WSI slide $\hat{Y}$ with only the ground truth bag label $Y$ supervision available. The bag label $Y$ would be positive if and only if there is at least one positive instance in the bag, and it is defined as \cref{eq:mil}:
\begin{equation}\label{eq:mil}
  Y = 
    \begin{cases}
      1 & \exists i, y_i=1,\\
      0 & \forall i, y_i=0,\\
    \end{cases}
\end{equation}
where $y_i$ is the ground truth label of the $i$-th patch. One issue with this formulation is that it cannot identify the discriminative instances, since each instance has the same weight under this framework. AB-MIL \cite{ilse_attention-based_2018} is thus a suitable solution. AB-MIL assigns attention scores to the instances, which reflect the importance of each instance in the final decision-making stage. Mathematically, assuming the hidden representation of the $i$-th instance is $\boldsymbol{h}_i \in \R^{1\times d} $ with the corresponding attention score $a_i$, then the bag-level representation $\boldsymbol{h}_{b}\in \R^{1\times d_0}$ is calculated through \cref{eq:abmil}:
\begin{equation}\label{eq:abmil}
    \boldsymbol{h}_{b} = \sum_{i=1}^{n} a_i\boldsymbol{h}_i.
\end{equation}
Under the AB-MIL framework, the characteristic of WSIs having multiple resolutions is neglected. In our work, we incorporate multiple resolutions of WSIs with AB-MIL and propose the HAG-MIL framework. The HAG-MIL framework is designed based on AB-MIL and its motivation is that the attention scores from lower resolutions can be used as a compass to guide the model at a higher resolution. The differences between AB-MIL and our HAG-MIL framework are illustrated in \cref{fig:HAGMIL_illustration}. The inputs of the HAG-MIL framework are the features from different resolutions, and after the model at the lowest resolution has assigned attention scores to the features, the features with low attention in the higher resolution are discarded in the model at a higher resolution. The process is repeated until the highest resolution is reached.

\subsection{Hierarchical Attention-Guided MIL}\label{sec:HAG_MIL}
The overview of our proposed HAG-MIL framework is illustrated in \cref{fig:archi}. Given the WSI resolutions level $l$ to level $0$, the instance representations $(\boldsymbol{F}^l, \cdots, \boldsymbol{F}^0)$ extracted from multiple resolutions of any given WSI slide and the ground truth bag label $Y$, the HAG-MIL framework learns a map from the instance representations $(\boldsymbol{F}^l, \cdots, \boldsymbol{F}^0)$ to the probability distribution of the bag label $(\hat{\boldsymbol{p}}^l, \cdots, \hat{\boldsymbol{p}}^0)$ and the attention scores of each instance $(\boldsymbol{a}^l, \cdots, \boldsymbol{a}^0)$.

Our HAG-MIL framework is defined in a recursive way. Given the instance representations $(\boldsymbol{F}^l, \cdots, \boldsymbol{F}^0)$, the IATs on different levels of resolutions $\mathcal{I} = \{\mathcal{I}_j|j=0,\cdots, l\}$, the HAG-MIL framework is given in  \cref{eq:hag_mil}:
\begin{equation}\label{eq:hag_mil}
\boldsymbol{a}^j, \hat{\boldsymbol{p}}^j = 
\begin{cases}
\begin{aligned}
    &\mathcal{I}_j \left(\boldsymbol{F}^j\right),\  \operatorname{if }j=l,\\
    &\mathcal{I}_{j} \left(\boldsymbol{F}^j_d\right), \operatorname{if }j\neq l,\\
\end{aligned}
\end{cases}
\end{equation}
where $\boldsymbol{F}^j_d$ is the distilled features at level $j$, defined in \cref{eq:distil_func}:
\begin{equation}\label{eq:distil_func}
    {\boldsymbol{F}_d^j} = \boldsymbol{F}^{j}\left[\operatorname{find\_topk\_ids}\left(\boldsymbol{a}^{j+1}, k_{j+1}\right)\right],
\end{equation}
where $k_{j+1}$ is the number of highest attention instances to be reserved at level $j+1$, and $\operatorname{find\_topk\_ids}(\boldsymbol{a}^{j+1}, k_{j+1})$ first finds the indices of the instances with top-$k_{j+1}$ attention scores at level $j+1$ and then generate the distilled indices of instances at level $j$, which contains \textbf{exactly the same content} as the instances at level $j+1$.

Finding the instances in different resolutions containing exactly the same content is a bit tricky, since patching and segmenting outcomes tend to be different in different resolutions, especially near the borderlines. For example, as shown in \cref{fig:archi}, patch $i$ at level 1 has four sub-patches at level 0, \ie, $4i,4i+1,4i+2,4i+3$; however, when we run the patching and segmenting algorithm at level 0 resolution, patches $4i+2$ and $4i+3$ will be dropped since they are backgrounds, while for the patch at level 1, patch $i$ as a whole is reserved. If some patches are dropped as stated before and some are not, since they are not near the borderlines, it is difficult to retrieve the sub-patches by indexing in higher resolutions.

To retrieve patches more effectively, a quadtree structure for each WSI is constructed. This structure utilises the hierarchical nature of WSIs, where one patch node has four child nodes that contain the exact same content.
As shown in \cref{fig:archi}, the patches are stored in the same sequence so that there is a clear map from an index in lower resolution to indices in higher resolutions, which is given in \cref{eq:index}:
\begin{equation}\label{eq:index}
    \operatorname{find\_sub\_patch\_ids}(i) = (4i, 4i+1, 4i+2, 4i+3).
\end{equation}

Then the find\_topk\_ids can be expressed as \cref{eq:find_top_k}:
\begin{equation}\label{eq:find_top_k}
\begin{aligned}
    &\operatorname{find\_topk\_ids}(\boldsymbol{a}^{j+1}, k_{j+1})\\
    =&\operatorname{find\_sub\_patch\_ids}(\operatorname{argsort}(\boldsymbol{a}^{j+1}, k_{j+1})).
\end{aligned}
\end{equation}

After obtaining the pair of bag label probability distribution $\hat{\boldsymbol{p}}^j$ and attention scores $\boldsymbol{a}^j$ at level $j$, the model at level $j$ is updated through back-propagation. Following the loss functions of \textbf{CL}ustering-constrained-\textbf{A}ttention \textbf{M}ultiple-instance learning (CLAM) \cite{lu_data-efficient_2021}, the loss function in our HAG-MIL framework is calculated through \cref{eq:loss}:
\begin{equation}\label{eq:loss}
    \mathcal{L}^j=\sum_{i=1}^N\operatorname{CE}\left(\hat{\boldsymbol{p}}^j, Y\right) + \lambda\operatorname{SS}\left(\hat{\boldsymbol{p}}_{p}^j, \boldsymbol{y}_{p}^j\right),
\end{equation}
where $\operatorname{CE}(\cdot,\cdot)$ and $\operatorname{SS}(\cdot,\cdot)$ stand for the \textbf{C}ross-\textbf{E}ntropy loss and the \textbf{S}mooth \textbf{S}upport Vector Machine loss \cite{Smoothed_top_k_loss_2018}, $\lambda$ is the balancing coefficient, $\hat{\boldsymbol{p}}_{p}^j\in \R^{k\times C}$ is the patch label probability distribution of the top-$k$ highest attention patches and $\boldsymbol{y}_{p}^j = [Y, \cdots, Y]^T\in \R^{k\times 1}$ are the generated labels of the top-$k$ highest attention patches. The patch-level loss is designed upon the idea that the patches with the highest attention scores should have the same label as the bag label.

\subsection{Integrated Attention Transformer}\label{sec:IAT}
The architecture of the proposed model together with the HAG-MIL framework is illustrated in \cref{fig:archi}. The vanilla transformer layer outputs only the contextual hidden representation of the input. In contrast, our IAM produces not only the contextualised hidden representation of the input but also the attention scores and the bag representation of the WSI slide. Furthermore, our IAM does not require a class token and comprises three constituents.

\textbf{Feedforward Neural Network (FNN)} The FNN in \cref{fig:archi} is a sequence of modules, a fully connected layer, a layer normalisation module, and a \textbf{G}aussian \textbf{E}rror \textbf{L}inear \textbf{U}nit (GELU) layer \cite{gelu_2016}. Given the input feature matrix $\boldsymbol{H}\in \R^{n\times d_{1}}$ and a learnable matrix $\boldsymbol{W} \in \R^{d_{1}\times d_2}$, the $\operatorname{FNN}(d_{1}, d_{2})$ is given in \cref{eq:FNN}:
\begin{equation}\label{eq:FNN}
    \operatorname{FNN}(d_{1}, d_{2})\left(\boldsymbol{H}\right) = \operatorname{GELU}\left(\operatorname{LayerNorm}\left(\boldsymbol{HW}\right)\right).
\end{equation}

\textbf{Transformer Layer} After the first FNN, we incorporate a transformer layer \cite{transformer_2017} to capture latent relationships between patches and provide cross-patch attention and additional contextual information to the original representations extracted by the pre-trained neural network. However, the time and space complexity of the vanilla transformer layer is $\bigoh(n^2)$, where $n$ is the number of patches and it is usually extremely huge for WSIs. To reduce time and memory costs, we employ Nystr\"{o}m attention \cite{Xiong_Zeng_Chakraborty_Tan_Fung_Li_Singh_2021} as our base transformer layer \cite{shao_transmil}, which has achieved competitive performance with the vanilla transformer layer and the time and space complexity are both $\bigoh(n)$. In addition, we also include the adaptive model initialization \cite{liu2020admin} to stabilise the training process.

\textbf{Aggregation Module} We use the gated attention mechanism \cite{ilse_attention-based_2018} as the aggregation module in our work, and we formulate it as a two-component module that can be easily extended in future research. 

The aggregation module can be divided into two components: (1) the attention component $s(\cdot):\R^{n\times d} \rightarrow \R^{n\times 1}$, which assigns attention scores $\boldsymbol{a} = s(\boldsymbol{H}) \in \R^{n\times 1}$ to the instances in the bag, and (2) the aggregation component $g(\cdot, \cdot):(\R^{n\times 1}, \R^{n\times d}) \rightarrow \R^{1\times d}$, which takes the attention scores $\boldsymbol{a}$ and the instance representations $\boldsymbol{H}$ as inputs and produces the bag representation $\boldsymbol{h}_{b}$. The aggregation module $f_a(\cdot): \R^{n\times d} \rightarrow \R^{1\times d}$ is the composition of the two components as described in \cref{eq:aggregation}:
\begin{equation}\label{eq:aggregation}
    \boldsymbol{h}_b = f_a(\boldsymbol{H}) = g(s(\boldsymbol{H}),\boldsymbol{H}).
\end{equation}
There are multiple options for the attention component $s(\cdot)$, including mean pooling, max pooling, the attention mechanism, the gated attention mechanism and so on. We propose using the gated attention mechanism \cite{ilse_attention-based_2018} as the aggregation module $s(\cdot)$ and it is shown in \cref{eq:attn_gated}:
\begin{equation}\label{eq:attn_gated}
    \boldsymbol{a}=s(\boldsymbol{H})
    =\left(\tanh \left(\boldsymbol{H}\boldsymbol{V}\right) \odot \sigma\left(\boldsymbol{H}\boldsymbol{U}\right)\right)\boldsymbol{w}_a,
\end{equation}
where $\boldsymbol{w}_a\in\R^{d_2\times 1}, \boldsymbol{V}, \boldsymbol{U}\in \R^{d_1\times d_2}$ are learnable matrices, $\boldsymbol{H} \in \R^{n\times d_1}$
is the hidden representation of the instances, $\sigma(\cdot)$ is the sigmoid function, $\operatorname{tanh(\cdot)}$ is the tanh function and $\odot$ stands for the element-wise production. 

The aggregation component $g(\cdot,\cdot)$ can be the weighted average on the instance representation based on the attention scores. The bag representation $\boldsymbol{h}_{b} \in \R^{1\times d}$ is given according to \cref{eq:abmil}. Therefore, $f_a(\cdot)$ in our paper is given by \cref{eq:weighted_avg}:
\begin{equation}\label{eq:weighted_avg}
        f_a(\boldsymbol{H}) = (\left(\tanh \left(\boldsymbol{H}\boldsymbol{V}\right) \odot \sigma\left(\boldsymbol{H}\boldsymbol{U}\right)\right)\boldsymbol{w}_a)^T \boldsymbol{H}.
\end{equation}
Finally, since an IAT consists of $m$ IAMs, it produces $m$ bag representations. Bag representations from different IAMs may capture different features of the bag; therefore, different bag representations should have different importance scores when aggregated into the final bag representation. Hence, we use an attention mechanism to fuse the bag representations from each IAM into the IAT bag representation. Assuming the bag representation from the $i$-th IAM is $\boldsymbol{h}_{b_i}$, and the learnable vector $\boldsymbol{w}_b\in \R^{m\times 1}$ with each element corresponding to the contribution of each $\boldsymbol{h}_{b_i}$, the final bag representation $\boldsymbol{h}_{bf} \in \R^{1\times d}$ is given in \cref{eq:final_slide}:
\begin{equation}\label{eq:final_slide}
    \boldsymbol{h}_{bf} = \boldsymbol{w}_b^T
    \begin{bmatrix}
        \boldsymbol{h}_{b_1}\\
        \vdots\\
        \boldsymbol{h}_{b_m}
    \end{bmatrix}.
\end{equation}
The last step is to feed the $\boldsymbol{h}_{bf}$ into a classification layer to obtain the bag-level distribution $\hat{\boldsymbol{p}}$.

\begin{table*}[t]
\begin{center}
\begin{tabular}{l|ccc|ccc}
\hline
\multirow{2}{*}{\bf METHOD} & \multicolumn{3}{c|}{\bf Camelyon16} & \multicolumn{3}{c}{\bf Gastric}\\
& \textbf{AUC}$\uparrow$ & \textbf{F1}$\uparrow$ & \textbf{Accuracy}$\uparrow$ &\textbf{AUC}$\uparrow$ & \textbf{F1}$\uparrow$ & \textbf{Accuracy}$\uparrow$\\
        \hline\hline
        Max Pooling & 0.715$_{0.007}$ & 0.549$_{0.012}$ & 0.682$_{0.005}$ & 0.871$_{0.007}$ & \sbest{0.601}$_{0.053}$ &\sbest{0.776}$_{0.071}$\\
        Mean Pooling & 0.654$_{0.075}$ & 0.437$_{0.035}$ & 0.640$_{0.013}$ & 0.839$_{0.011}$ & 0.551$_{0.014}$ & 0.704$_{0.020}$\\
        AB-MIL~\cite{ilse_attention-based_2018} & 0.903$_{0.013}$ & 0.850$_{0.018}$ & 0.868$_{0.015}$ & \sbest{0.886}$_{0.016}$ & 0.582$_{0.028}$ & 0.743$_{0.038}$\\
        CLAM-SB~\cite{lu_data-efficient_2021} &0.834$_{0.068}$ & 0.782$_{0.059}$ & 0.819$_{0.043}$ & 0.838$_{0.045}$ & 0.533$_{0.063}$ & 0.673$_{0.091}$\\
        CLAM-MB~\cite{lu_data-efficient_2021} & 0.876$_{0.024}$ & 0.800$_{0.031}$ &  0.826$_{0.019}$ & 0.873$_{0.018}$ & 0.578$_{0.023}$ & 0.742$_{0.033}$ \\
        TransMIL~\cite{shao_transmil} & 0.838$_{0.047}$ & 0.787$_{0.033}$ & 0.809$_{0.027}$ & 0.834$_{0.016}$ & 0.560$_{0.046}$ & 0.731$_{0.082}$\\
        DTFD-MIL~\cite{zhang_dtfd-mil_2022} & \sbest{0.933}$_{0.009}$ & \sbest{0.858}$_{0.017}$ & \best{0.899}$_{0.012}$ & \best{0.890}$_{0.011}$ & 0.565$_{0.020}$ & 0.725$_{0.030}$\\
        \hline\hline
        \rowcolor{LightCyan}
        HAG-MIL (Ours) & \best{0.946}$_{0.011}$ & \best{0.874}$_{0.025}$ & \sbest{0.887}$_{0.022}$ & \best{0.890}$_{0.011}$ & \best{0.616}$_{0.027}$ & \best{0.781}$_{0.027}$\\
        \hline\hline
        \rowcolor{LightCyan}
        Improvement & 0.013 & 0.016 & -0.012 & 0.000 & 0.015 & 0.005\\
        \hline
\end{tabular}
\caption{Results on Camelyon16 and our gastric cancer test set. The best ones are in red bold, and the second best ones are in blue italic underline. The subscript in each cell is the standard derivation.}
\label{tab:came_gastric}
\end{center}
\end{table*}
\begin{table*}[t]
\begin{center}
\begin{tabular}{l|ccc|ccc}
\hline
\multirow{2}{*}{\bf METHOD} & \multicolumn{3}{c|}{\bf TCGA-RCC} & \multicolumn{3}{c}{\bf TCGA-NSCLC}\\
& \textbf{AUC}$\uparrow$ & \textbf{F1}$\uparrow$ & \textbf{Accuracy}$\uparrow$ &\textbf{AUC}$\uparrow$ & \textbf{F1}$\uparrow$ & \textbf{Accuracy}$\uparrow$\\
        \hline \hline
        Max Pooling & 0.970$_{0.000}$ & 0.847$_{0.008}$ & 0.880$_{0.005}$ & 0.863$_{0.001}$ & 0.774$_{0.002}$ & 0.774$_{0.002}$\\
        Mean Pooling & 0.959$_{0.010}$ & 0.828$_{0.015}$ & 0.862$_{0.014}$ & 0.851$_{0.000}$ & 0.777$_{0.002}$ & 0.777$_{0.002}$\\
        AB-MIL~\cite{ilse_attention-based_2018} & \sbest{0.978}$_{0.002}$ & 0.864$_{0.013}$ & 0.892$_{0.010}$ & 0.903$_{0.007}$ & 0.817$_{0.010}$ & 0.817$_{0.010}$\\
        CLAM-SB~\cite{lu_data-efficient_2021} & 0.970$_{0.007}$ & 0.846$_{0.011}$ & 0.878$_{0.010}$ & \sbest{0.905}$_{0.004}$ & 0.824$_{0.007}$ & 0.824$_{0.007}$\\
        CLAM-MB~\cite{lu_data-efficient_2021} & 0.975$_{0.005}$ & 0.871$_{0.012}$ & 0.895$_{0.010}$ & 0.900$_{0.004}$ & 0.818$_{0.010}$ & 0.818$_{0.010}$ \\
        TransMIL~\cite{shao_transmil} & 0.972$_{0.003}$ & 0.850$_{0.007}$ & 0.876$_{0.011}$ & 0.902$_{0.008}$ & 0.825$_{0.015}$ & 0.825$_{0.016}$\\
        DTFD-MIL~\cite{zhang_dtfd-mil_2022} & 0.976$_{0.009}$ & \sbest{0.884}$_{0.028}$ & \sbest{0.898}$_{0.027}$ & 0.902$_{0.023}$ & \sbest{0.830}$_{0.027}$ & \sbest{0.838}$_{0.030}$\\
        \hline \hline
        \rowcolor{LightCyan}
        HAG-MIL (Ours) & \best{0.982}$_{0.001}$ & \best{0.894}$_{0.009}$ & \best{0.914}$_{0.006}$ & \best{0.921}$_{0.008}$ & \best{0.849}$_{0.015}$ & \best{0.849}$_{0.016}$\\
        \hline\hline
        \rowcolor{LightCyan}
        Improvement & 0.004 & 0.010 & 0.016 & 0.016 & 0.019 & 0.011\\
        \hline
\end{tabular}
\caption{Results on TCGA-RCC and TCGA-NSCLC test set.}
\label{tab:rcc_nsclc}
\end{center}
\end{table*}

\section{Experiments and Results}
We first compared our method with other methods on various datasets. Then we conducted extensive sensitivity and ablation studies to validate the effectiveness of the HAG-MIL framework and the aggregation module. Finally, we visualised the attention scores from our method and compared them with those from the CLAM model.
\subsection{Dataset Descriptions}
\textbf{Camelyon16 Dataset} \cite{litjens_1399_2018} It contains 397 WSIs of lymph nodes with and without metastasis in tissue sections of women with breast cancer. The training dataset includes 157 normal WSIs and 111 tumour WSIs. The test dataset is provided separately and contains 129 test WSIs, including both tumour and normal WSIs. We further split the training dataset into a new training and validation dataset by 7:3 and evaluate all of the methods on the official test dataset. 

\textbf{IMGC Dataset} The \textbf{I}ntestinal \textbf{M}etaplasia \textbf{G}astric \textbf{C}ancer (IMGC) dataset is an in-house dataset consisting of 1,882 WSIs, obtained from the Periodic Acid-Schiff-Alcian Blue stained biopsies of the patients diagnosed with intestinal metaplasia. The patients were followed to determine whether they ultimately developed gastric cancer. We used only a portion of this dataset due to its extreme imbalance. The training dataset includes 130 normal and 43 cancer cases, the validation dataset includes 34 normal and 13 cancer cases, and the test dataset includes 253 normal and 18 cancer cases.

\textbf{TCGA-RCC Dataset} It includes 940 WSIs in total. Individually, project Kidney Chromophobe Renal Cell Carcinoma (TGCA-KICH) has 121 WSIs from 109 cases. Project Kidney Renal Clear Cell Carcinoma (TCGA-KIRC) has 519 WSIs from 513 cases, and project Kidney Renal Papillary Cell Carcinoma (TCGA-KIRP) has 300 WSIs from 276 cases. We split the dataset into training, validation, and test datasets by the ratio of 6:1.5:2.5, respectively.

\textbf{TCGA-NSCLC Dataset} It includes 1,053 WSIs in total. Individually, project Lung Squamous Cell Carcinoma (TCGA-LUSC) has 512 WSIs from 478 cases, and project Lung Adenocarcinoma (TCGA-LUAD) has 541 WSIs from 478 cases. We split the dataset into training, validation, and test datasets by the ratio of 6:1.5:2.5, respectively. 

\subsection{Implementation Details}
\textbf{Evaluation Metrics} The evaluation criteria are \textbf{A}rea \textbf{U}nder the \textbf{C}urve (AUC), F1, and accuracy scores. The AUC score reflects the overall ability to distinguish different classes. The F1 score measures the true performance when the data is imbalanced. Together, these metrics can holistically assess the model performances. The thresholds for F1 and accuracy scores are determined by Youden's index in the IMGC dataset due to its extreme unbalance and they are set to 0.5 for other datasets, following previous works.

\textbf{Training Settings} Baseline methods include max/mean pooling, AB-MIL~\cite{ilse_attention-based_2018}, CLAM-SB, CLAM-MB~\cite{lu_data-efficient_2021}, TransMIL~\cite{shao_transmil}, and DTFD-MIL~\cite{zhang_dtfd-mil_2022}. Training is terminated after 20 epochs of no decrease in validation loss. Official implementations of the methods were used when available, every effort was made to reproduce the results. For the IMGC dataset, all methods were run at level 0 resolution. For other datasets, baseline methods were run at level 1 resolution, following their respective works. The models were run five times and the average performances were reported.

\textbf{Hyper-parameters Details}
The learning rate and weight decay are set to 10$^{\text{-5}}$. Dropout is set to 0.25 for the gated attention mechanism and 0.3 for the transformer layer. The Swin-Transformer \cite{swin_transformer} large is used to extract the features from the IMGC dataset. ResNet-50 \cite{resnet_he_16} is used for other datasets, following previous works. Therefore, the input dimension for the IMGC dataset and other datasets are 1,536 and 1,024, respectively. The dimensions of the stacked IAMs are determined via experimental validation. Interestingly, our experiment found that the optimal structure for the model consisted of four modules with dimensions of 1,024, 1,536, 512, and 1,024, respectively, forming a bottleneck structure. The dimension of the final bag representation is 1,024. The loss functions for models at different resolutions are different because we tolerate slight errors in lower resolutions, but the model at the highest resolution must be as accurate as possible. Therefore, label smoothing is set to 0.1 for low resolutions, and is not used for the highest resolution. The balancing coefficient $\lambda$ is set to 1. 

\subsection{Comparison with State-of-the-art Methods} \Cref{tab:came_gastric} and \Cref{tab:rcc_nsclc} show all results of the experiments. Our method had the highest AUC and F1 scores and the highest or competitive accuracy scores across all datasets, which proved the effectiveness of all our proposed methods.

Camelyon16 is quite challenging due to the small area of tumour regions. From \cref{tab:came_gastric}, max/mean pooling can barely learn the distinctions between tumour and normal slides, and the CLAM and TransMIL are inferior to our method by at least 4.3\% on the AUC score. Our method achieved a 1.3\% improvement in AUC score and a 1.6\% improvement in F1 score compared to DTFD-MIL, demonstrating its efficacy.

For the IMGC dataset, the fact that our method has achieved the best performance has proved that our proposed IAT is superior to other methods. Furthermore, the data imbalance problem on the IMGC dataset is much more severe than the other datasets, as the number of patients who developed gastric cancer is much less than those who did not. Hence, the result proved that our model performs well on extremely imbalanced datasets as well.

For the TCGA-RCC and TCGA-NSCLC datasets, they share a common characteristic that the tumour regions are large, which sometimes takes around 80\% of the whole WSI, making most of the methods to have similar performance, and even the most naive approaches like mean pooling and max pooling have achieved very high AUC scores. Our method has outperformed all others in every metric on these two datasets by a significant margin, particularly on the TCGA-NSCLC dataset. In addition, the highest AUC score of our method on the TCGA-RCC dataset demonstrated that our model also excels in the multi-class classification task. 

\begin{figure}
    \centering
    \includegraphics[width=\linewidth]{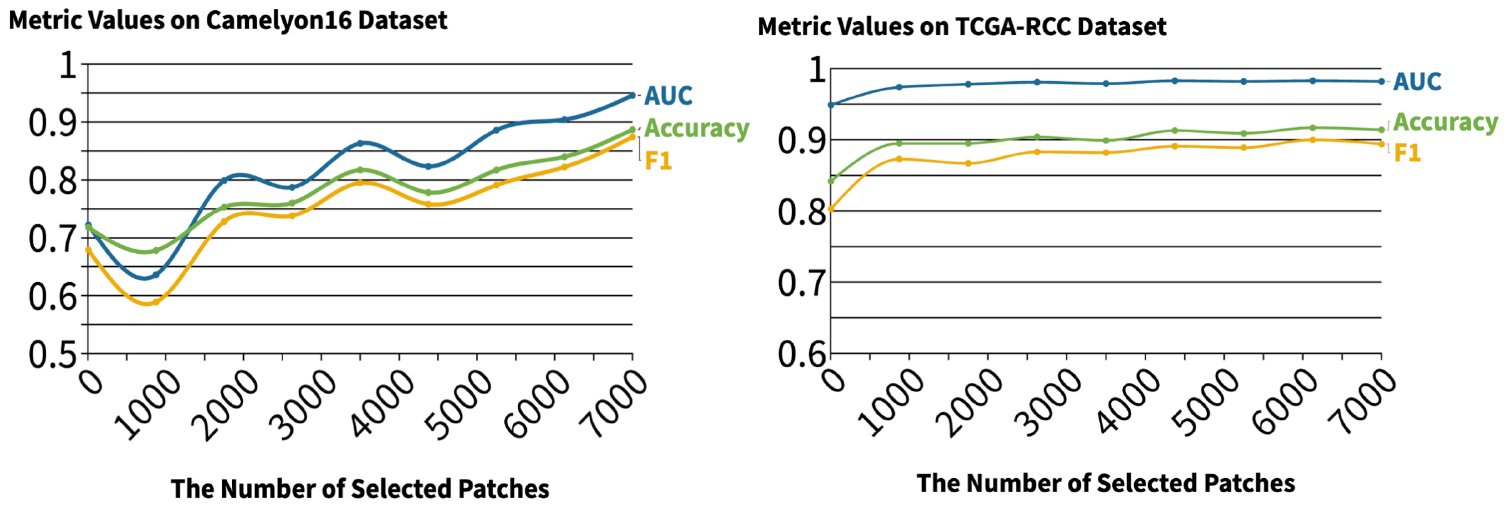}
    \caption{AUC, F1 and accuracy scores with different $k$ values on the Camelyon16 (left) and TCGA-RCC (right) datasets.}
    \label{fig:ablation}
\end{figure}

\begin{table}[t]
\begin{center}
\begin{tabular}{l|ccc}
\hline
\multirow{2}{*}{\bf METHOD} & \multicolumn{3}{c}{\bf Camelyon16}\\
  & \textbf{AUC}$\uparrow$ & \textbf{F1}$\uparrow$ & \textbf{Accuracy}$\uparrow$ \\
        \hline
        CLS w/o gated & 0.917$_{0.040}$ & 0.834$_{0.026}$ & 0.855$_{0.020}$\\
        CLS w/ gated & \sbest{0.929}$_{0.020}$ & \sbest{0.855}$_{0.044}$ & \sbest{0.869}$_{0.038}$\\
        Aggr w/o gated & 0.928$_{0.012}$ & 0.848$_{0.040}$ & 0.863$_{0.038}$\\
        \hline\hline
        \rowcolor{LightCyan}
        Aggr w/ gated & \best{0.946}$_{0.011}$ & \best{0.874}$_{0.025}$ & \best{0.887}$_{0.022}$\\
        \hline
\end{tabular}
\caption{The results on the use of several aggregation modules and the class token on the Camelyon16 dataset.}
\label{tab:aggre_cls}
\end{center}
\end{table}

\subsection{Sensitivity and Ablation Studies}\label{sec:fine_tune}
We performed sensitivity and ablation analyses to assess the impact of our HAG-MIL framework and IAT on performance.

\textbf{The HAG-MIL Framework} The number of patches ($k_j$) reserved for higher resolution is a crucial parameter in our framework. $k_j$ should not be too small to avoid missing tumour patches, but it should also not be too large to avoid time and GPU memory issues. Since tumour regions are small in Camelyon16 and large in the TCGA series dataset, they tend to have different trends with changes of $k_j$; therefore, we conducted experiments on both Camelyon16 and TCGA-RCC datasets. For convenience, we set $k_1=k_2$. Eight different $k_j$ values are sampled within the range of [0, 7,000] at equal intervals, with the starting point of $k_j=\text{7}$. In general, larger values of $k_j$ tend to result in improved model performance. However, the extent of this influence varies across different datasets. The $k_j$ value heavily influences how the model performs on Camelyon16 due to the small tumour regions. Our findings also indicate that increasing the value of $k_j$ during the inference stage can further enhance model performance. However, the $k_j$ value does not influence the performance on the TCGA-RCC dataset much because the tumour regions are large, and there is a high probability that the true discriminative patches are in the top-$k_j$ patches even when $k_j=\text{7}$.

\textbf{Aggregation Module} We compared transformers with and without an aggregation module and included two types of aggregation modules in our experiments: the attention mechanism \cite{ilse_attention-based_2018} and the gated attention mechanism \cite{ilse_attention-based_2018}. In order for a fair comparison, the patch-level loss is also included in transformers using the class token, and the attention scores used by our HAG-MIL framework are generated by an aggregation module in the final layer. The bag representations are still produced by the class token. The results are shown in \cref{tab:aggre_cls}. From these results, we can draw two conclusions: (1) the gated attention mechanism is a better option for the aggregation module as it outperforms the attention mechanism transformer by 1.8\% and 1.2\% in terms of the AUC score, and (2) our transformer with both two aggregation modules outperform the class token transformer by 1.7\% and 1.1\% in terms of the AUC score.
\begin{figure}
    \centering
    \includegraphics[width=\linewidth]{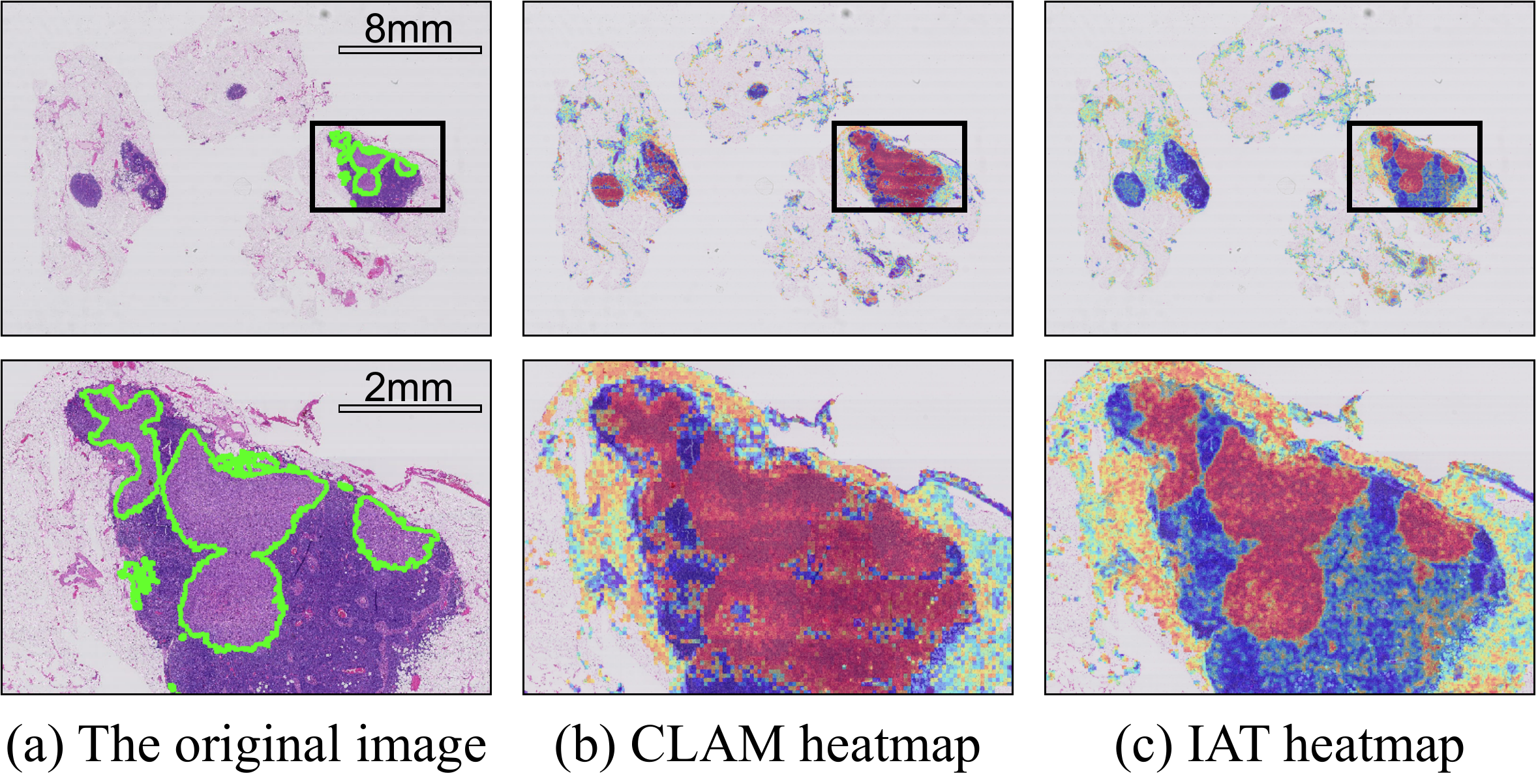}
    \caption{Figures depicting a WSI from Camelyon16 with ground truth annotations (a), alongside heatmaps generated by CLAM-SB (b) and IAT (c). The second row of figures displays magnified views of the regions enclosed by black boxes in the first row.}
    \label{fig:heatmap}
\end{figure}
\subsection{Visualisation of the Attention Scores}
We demonstrate the interpretability of our method by visualising the attention scores as heatmaps, which indicate the locations of the discriminative regions. For this experiment, we used the Camelyon16 dataset with ground truth pixel-level annotations and compared the heatmaps of our method with those of the CLAM-SB model. The original image and visualisation results are shown in \cref{fig:heatmap}. The ground truth tumour regions are bounded by green lines and the red regions in the heatmaps indicate the model-predicted high-attention regions. While the CLAM-SB model can capture most of the macro-metastasis in the WSI, it has a much higher false positive rate than our method. Additionally, the CLAM-SB model performs poorly at identifying micro-metastasis and isolated tumour cells due to their small size. However, our method accurately predicts macro-metastasis and can also localise micro-metastasis and isolated tumour cells with relatively smaller attention scores compared to macro-metastasis. It is clear that heatmaps from our method have better contrast and higher consistency with ground truth annotations.

\section{Conclusion}
In this paper, we present the HAG-MIL framework and the IAT for WSI classification tasks. The HAG-MIL framework fully utilises the WSIs by exploiting multiple resolutions of the WSIs, and it enables the large models, like our IAT, to run at the highest resolution of the WSIs. Diagnosing like a human pathologist, the HAG-MIL framework is capable of dynamically localising the most discriminative instances from the enormous number of instances in the WSI bag across multiple resolutions in a hierarchical and attentive way. Within the HAG-MIL framework, the IAT takes into account the evolvement of the bag representations through different transformer layers. It produces a more holistic and effective bag representation than the vanilla transformer by aggregating multiple bag representations generated by the IAMs. The experimental results show that our method outperforms other methods and obtains a more accurate heatmap. Our method can be further extended to more applications such as super-resolution image analysis, natural language processing, \etc.
\newpage
\section*{Acknowledgments}
The work described here was partially supported by grants from the National Key Research and Development Program of China (No. 2018AAA0100204) and from the Research Grants Council of the Hong Kong Special Administrative Region, China (CUHK 14222922, RGC GRF, No. 2151185). The results shown here are based upon data generated by the TCGA Research Network: \url{https://www.cancer.gov/tcga}.
\bibliographystyle{named}
\bibliography{ijcai23}
\end{document}